\documentclass{article}

\PassOptionsToPackage{numbers, compress}{natbib}

 \usepackage[final]{neurips_2021}

\usepackage[tableposition=top]{caption}
\usepackage[utf8]{inputenc} 
\usepackage[T1]{fontenc}    
\usepackage{hyperref}       
\usepackage{url}            
\usepackage{booktabs}       
\usepackage{amsfonts}       
\usepackage{nicefrac}       
\usepackage{microtype}      
\usepackage{xcolor}         
\usepackage{graphicx}
\usepackage{upgreek}
\usepackage{times}
\usepackage{epsfig}
\usepackage{amsmath}
\usepackage{amssymb}
\usepackage{comment}

\usepackage{array}
\newcolumntype{P}[1]{>{\centering\arraybackslash}p{#1}}
\newcolumntype{M}[1]{>{\centering\arraybackslash}m{#1}}

\title{XCI-Sketch: Extraction of Color Information from Images for \\ Generation of Colored Outlines and Sketches}

\author{V Manushree
\thanks{
Authors have contributed equally to this work and share first authorship. Link of the code: \href{https://github.com/Sampai28/GeneratedSketches}{https://github.com/Sampai28/GeneratedSketches}
}  
\quad Sameer Saxena$^{*}$ 
\quad Parna Chowdhury$^{*}$ 
\quad Manisimha Varma$^{*}$ 
\\
\textbf{
 Harsh Rathod$^{*}$
\quad Ankita Ghosh
\quad Sahil Khose
}
\\
\\
Manipal Institute of Technology, Manipal \\

{\tt\small \{
manushree.v, 
sameer.saxena,
parna.chowdhury, 
manthena.varma,
}\\
{\tt\small 
harsh.rathod, 
ankita.ghosh1, 
sahil.khose
\}@learner.manipal.edu}
}

\begin{document}

\maketitle
\begin{abstract}
  Sketches are a medium to convey a visual scene from an individual's creative perspective. The addition of color substantially enhances the overall expressivity of a sketch. This paper proposes two methods to mimic human-drawn colored sketches by utilizing the Contour Drawing Dataset. Our first approach renders colored outline sketches by applying image processing techniques aided by k-means color clustering. The second method uses a generative adversarial network to develop a model that can generate colored sketches from previously unobserved images. We assess the results obtained through quantitative and qualitative evaluations.
\end{abstract}

\section{Introduction}
\label{section:introduction}
Sketches are elementary representations of visual scenes. They are easier to draw and are not limited to artists, making them an accessible and common form of art. Sketches find application in various industries pertaining to advertisement, fashion designing, architecture, and interior designing, where illustrations play a significant role. Often while using grayscale sketches, the overall expressivity falls short, and it is challenging to derive the necessary information. Color, being an integral part of art, appeals to the human senses. It elevates the sketch and efficiently carries across its meaning, background details, and connotation. 

In recent years, image processing and deep learning techniques have been extensively deployed to synthesize artwork of various forms. Previously, attempts have been made to leverage images to generate sketches, but these sketches have been limited to black and white shades. Our paper focuses on different approaches to convert photographic pictures into colored sketches. We use the Contour Drawing Dataset presented in the work by Li \cite{photosketch_cmu} and propose two ways to extract color information from the images and amalgamate it with the corresponding sketches. Our three-fold contribution which is explained in detail in the succeeding sections of the paper is as follows: first, we formulate a process to transfer color onto the existing black and white sketches in the dataset to produce colored outlines. Second, we propose a method to produce colored sketches by performing colorspace manipulation. Finally, we go a step further to use these sketches as the training dataset for a generative adversarial network and develop a model which can produce colored sketches from unseen images.

We discuss the related literature based on which we have modeled our approaches in Section \ref{section:related works}. The algorithm for rendering colored outlines is explained in Section \ref{section:colored outline}. Section \ref{section:GANs sketch} describes the image processing techniques employed to produce color sketches, followed by the methodology used for generating colored sketches using a generative model. Section \ref{section:results} evaluates our results by comparing the produced sketches with existing sketch datasets and conducting a perceptual user study. We conclude by suggesting various possible applications of our work in Section \ref{section:applications} and the scope of improvement in Section \ref{section:conclusion}.


\section{Related Works}
\label{section:related works}

\paragraph{Approaches for rendering sketches:}A wide range of research work has been done by the computer vision and computer graphics community in the domain of sketch synthesis. Prior techniques include rendering sketches by applying edge detection \cite{canny,boundary}, contour detection \cite{photosketch_cmu} and image segmentation \cite{boundary_detection_2, boundary_detection_1}. Many sizeable sketch datasets have been collected through crowdsourcing by designing interactive user interfaces \cite{sketchsoup,quickdraw,Draw_Assistant}. The onset of deep learning substantially enhanced the results produced in this field. SketchRNN \cite{quickdraw}, a RNN-based deep variational autoencoder, was among the earlier works in deep learning that generated diverse sketches. Style transfer using neural network \cite{nst1,Li_and_Wang} was a revolutionary step in the generation of art by producing images that preserve content and style components of the input images. Neural style transfer has since been adopted extensively to produce sketches \cite{nstsketch1,colortransform}. Other notable works include the use of reinforcement learning \cite{reinforcement} to generate ink paintings from pictures and photographs and implementation of hidden Markov model \cite{HMM} to transform a coarse sketch into a refined drawing. Generative adversarial networks (GANs) have found immense application in developing sketch generation models.

\paragraph{GANs for generating sketches:}Generative adversarial network based models have seen major advancements in recent times, and this has motivated great research in its use as an art and image creation tool. Pencil-shading sketches have been produced by architectures like ArtPDGAN \cite{artpdgan} which combines an image-to-image network to produce a key map that aids the generator. BP-GAN \cite{bpgan}, and Composition Aided GANs \cite{cagan} perform face sketch synthesis using methods like back projection and compositional reconstruction loss, respectively. GANs have been used for colorizing anime sketches by applying conditional generation \cite{cgan}, and style transfer \cite{acgan}. DoodlerGAN \cite{creativegan} generates doodle-like sketches in a sequential drawing manner. RoboCoDraw \cite{robocodraw} is a real-time robot-based drawing system that stylizes human face sketches interactively and generates cartoon avatar face sketch from an individual's image using AvatarGAN. CartoonGAN \cite{cartoongan} proposes an architecture to convert real-world scenes into cartoon-style images by introducing semantic content loss and edge-promoting adversarial loss. Several GAN architectures have been used to retrieve images from sketches \cite{sketch_gan,deep_anime_gan,sbir,context_gan,scribbler,prog-gan}, and even generate three-dimensional models from sketches \cite{sketchsoup,lazy}. The ability of GANs to generate volumes of synthetic data makes it an ideal tool for expanding on existing datasets and creating new ones. 

\paragraph{Sketch Datasets:}Art generation is a broadly investigated domain in computer vision, which in turn gives rise to various applications for the dataset. Existing datasets like TU-Berlin \cite{tu-berlin} consists of $20000$ unique human-drawn sketches distributed over $250$ distinct object classes, which have been used to train classification models. The Sketchy Database \cite{sketchydatabase} comprising of $12500$ photographs and approximately $75000$ human drawn sketches inspired by the same, has been used to generate more photo-sketch pairs. SketchyScene \cite{SketchyScene} contains more than $29000$ scene-level sketches, and all objects in the sketches have ground-truth semantic and instance masks. This enables several applications like sketch colorization, editing, and captioning. QuickDraw \cite{quickdraw}, Sketch Me That Shoe \cite{SketchShoe} and CHUK Face Sketch Database \cite{FaceSketch} are examples of datasets that have been used for sketch generation, sketch-based image retrieval, and sketch synthesis and recognition, respectively.

The Contour Drawing \cite{photosketch_cmu} Dataset consists of $1000$ outdoor images, which are paired with five different versions of sketches, thus making a total of $5000$ human-drawn black and white sketches. These five sketches are presented in three variations of stroke width $1$, $3$, and $5$ units. Majority of the present sketch datasets contain grayscale sketches. We extend the Contour Drawing Dataset by rendering colored outlines and color-filled versions of the available sketches, which can be utilized for various tasks.

\section{Rendering Colored Outlines in Sketches}
\label{section:colored outline}
In this section, we discuss how we transfer color to the black and white sketches of the Contour Drawing Dataset. Our methodology incorporates several techniques, including k-means color clustering and bitwise operations. We also describe the method that we deploy to find the optimal number of clusters per image for the k-means algorithm. 
\begin{figure*}[t]
    \centering
    \includegraphics[scale=0.55]{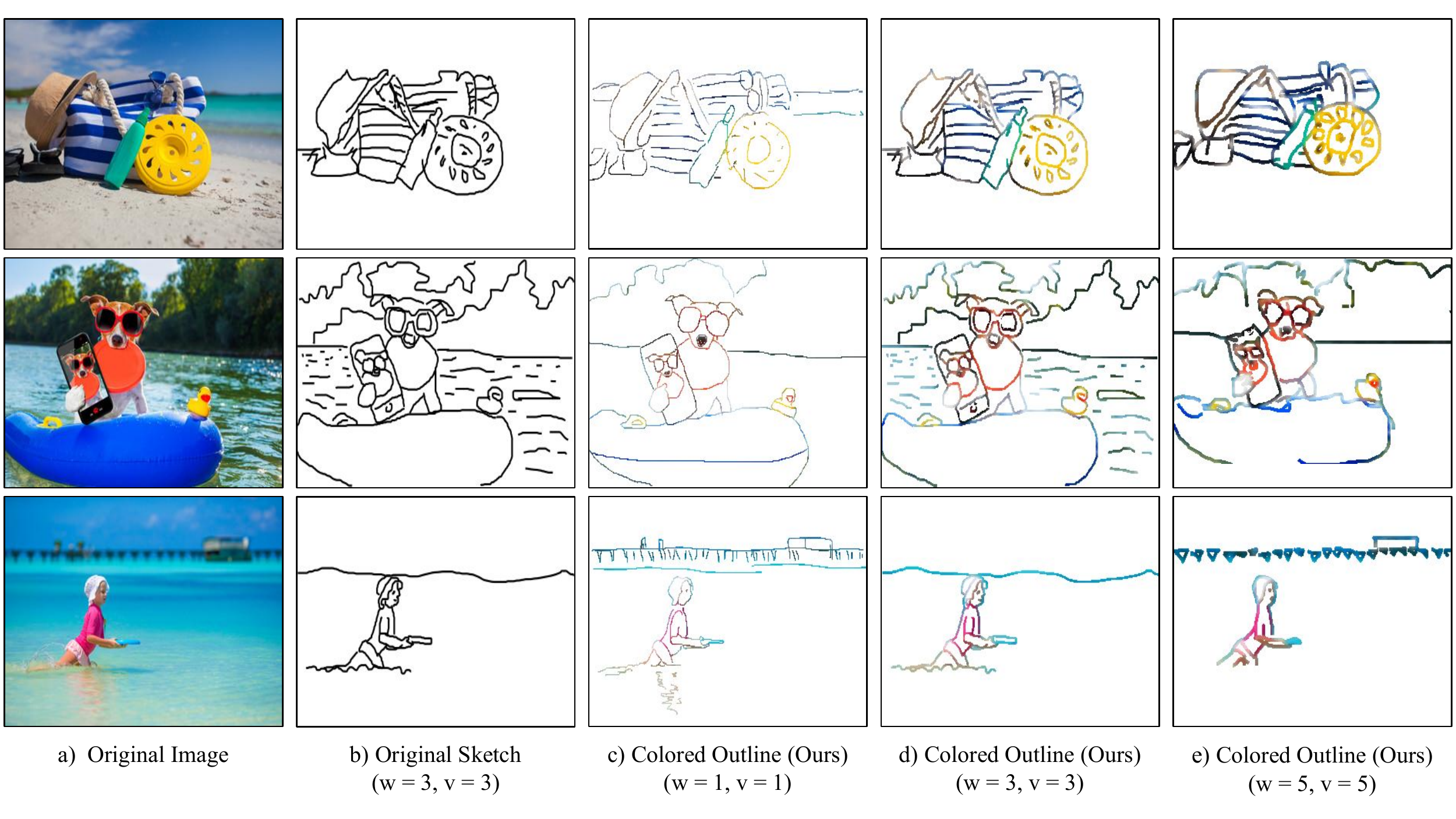}
    \caption{Qualitative results of the rendered colored outlines. (b) shows the sketch presented in \cite{photosketch_cmu} dataset. (c), (d) and (e) display the results obtained by us after color extraction from image (a). $w$ is the stroke width and $v$ is the version of sketch.}
    \label{fig:outlines_colored}
\end{figure*}



\subsection{Methodology}
A digital RGB image can contain $256\times256\times256$ different colors, but generally, only a few colors are used when an individual draws a sketch. To mimic this while generating colored outlines, we perform quantization \cite{color_quantization} of the number of colors in the images. First, we perform three iterations of Gaussian blur \cite{gaussian_blur} with a kernel size of $5\times5$ on the images. This is done to reduce the sharp color transitions inside the images as sketches are boundary-like drawings that capture the outline of the visual scene. To achieve quantization, we perform k-means clustering \cite{kmeans} on the image where the pixels are categorized into $k$ clusters according to their intensity value. We formulate an algorithm to determine this value $k$, which is described in Section \ref{subsection:kvalue}. 

Since the dimensions of the image and its corresponding sketches do not align in some cases, we slice the extraneous rows or columns to attain uniform dimensions throughout the dataset. We apply binary thresholding on the sketch and extract the black outlines to form a mask. We split the channels of the post-processed image and perform bitwise operations between the mask and each channel to obtain the color information. The resultant channels are merged to render the final colored outline sketch. 
\begin{figure}[t]
    \centering
    \includegraphics[scale=0.328]{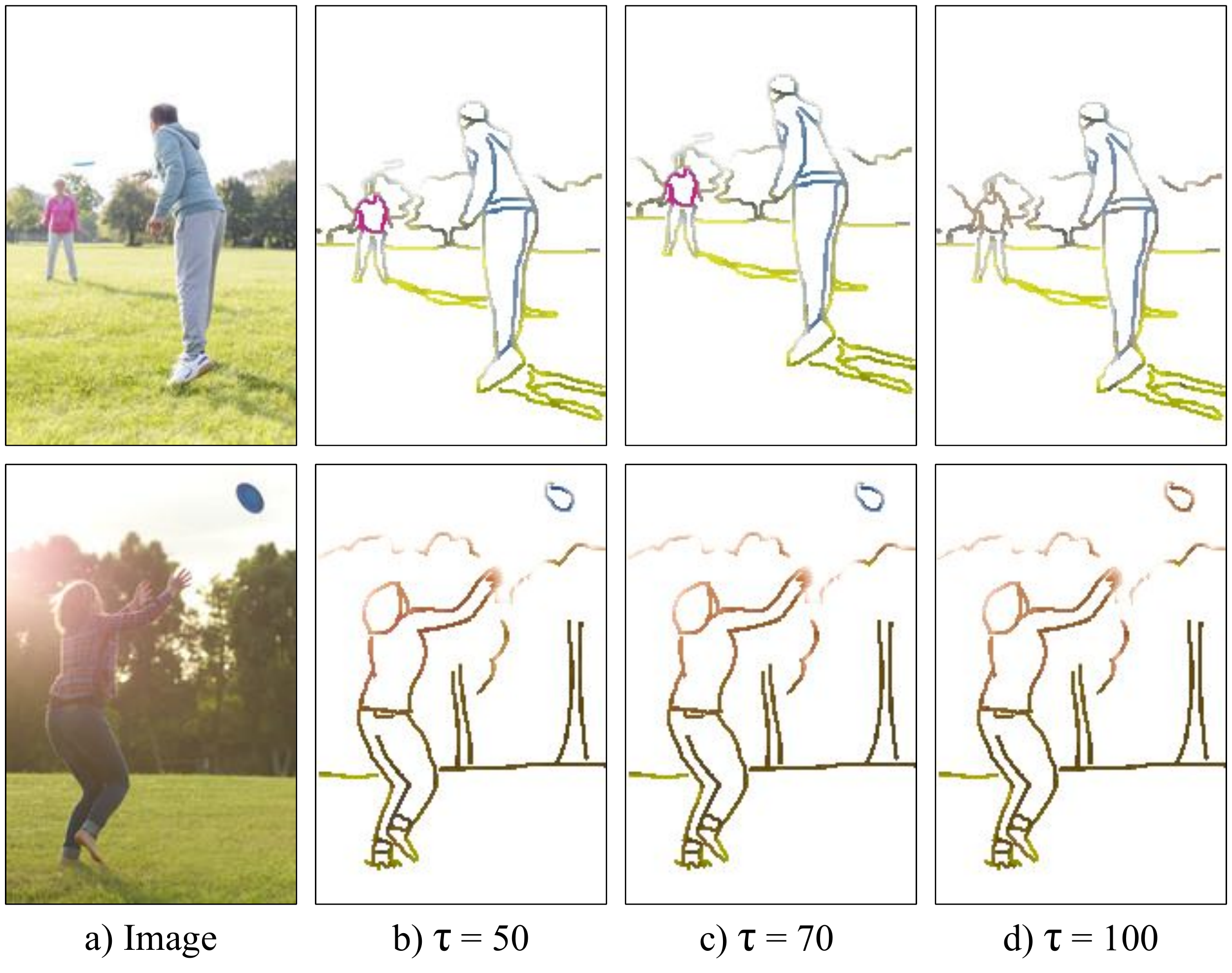}
    \caption{Colored outlines based on different threshold values. We can observe that $\uptau$ = $70$ is the optimal value. At higher threshold values ($\uptau$ = $100$), we can observe the loss of color information. At lower threshold values ($\uptau$ = $50$), the result is similar to that of the optimal value ($\uptau$ = $70$) but is computationally expensive. $\uptau$ denotes the threshold value.}
    \label{fig:comparision_thresh}
\end{figure}
We repeat this process for all $15$ versions of sketches obtained from an image. Figure \ref{fig:outlines_colored} shows our results on three variations of the black and white sketches. $w$ denotes the width of the sketch outline, and $v$ denotes the version of the sketch.

\subsection{Calculating Optimal Number of Color Clusters}
\label{subsection:kvalue}
The appropriate number of clusters into which the data should be grouped is crucial in an unsupervised technique. In k-means clustering, the elbow method and the silhouette method \cite{elbow_method} are among the broadly used algorithms to determine the optimal value of $k$. These methods prove ineffective for our objective as iterating through all $k$ values to reach the optimal value is computationally expensive, and we desire to capture only the most prominent colors in the image. Hence we devised an alternative method. As the k-means algorithm clusters data by minimizing the sum of squared distance within the cluster, we appoint this criterion, called inertia, as a threshold value. Inertia can be calculated as:
\begin{equation}
\label{eqn:kmeanscost}
    \sum_{i=0}^{n}\left \|c_{i}-x_{i} \right \|^{2}
\end{equation}
where $x$ denotes pixels, $x_{i}$ denotes value of $i$th pixel and $c_{i}$ denotes value of cluster centroid closest to $x_{i}$.
\begin{figure}[b]
    \centering
    \includegraphics[scale=0.33]{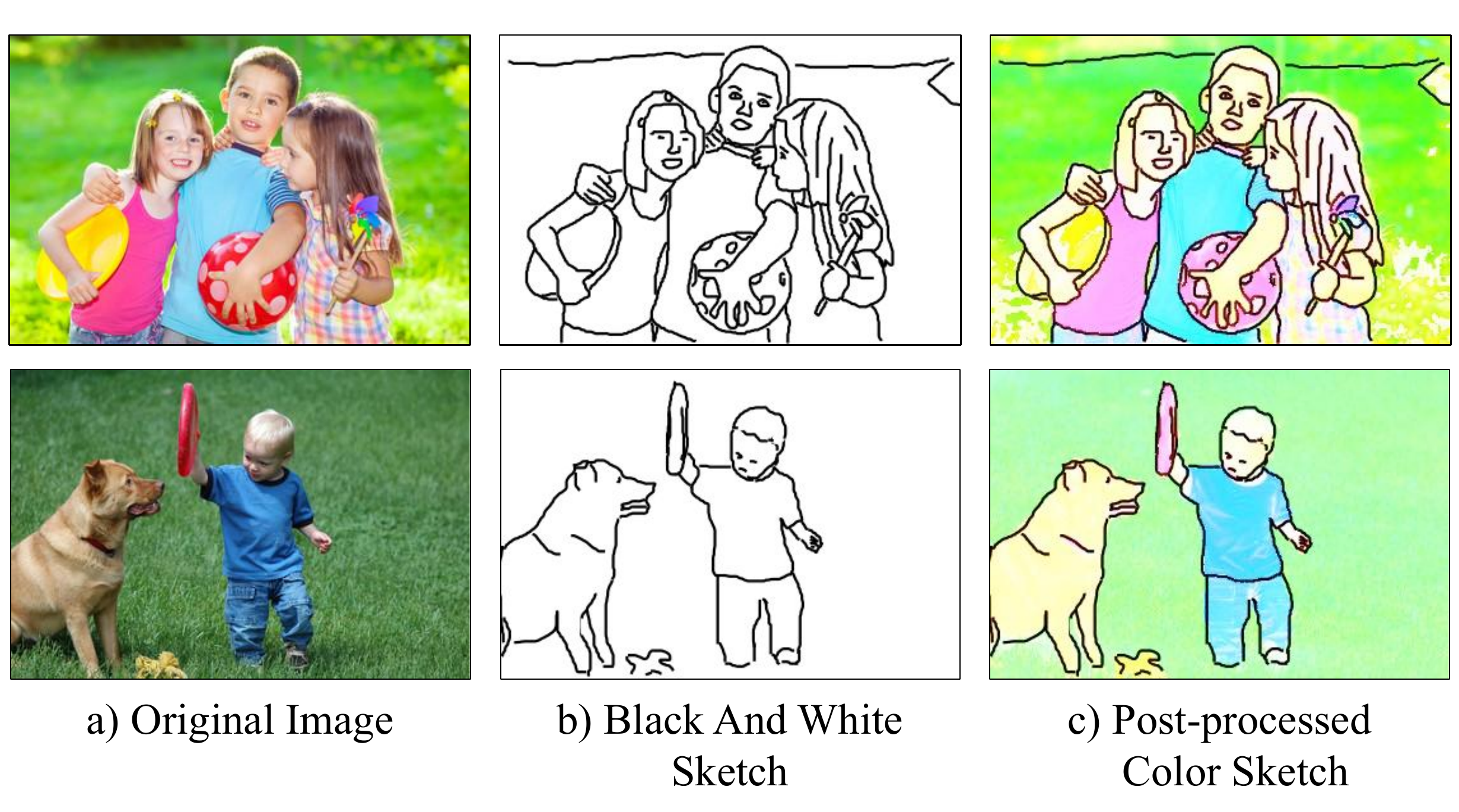}
    \caption{Qualitative results obtained after data preprocessing}
    \label{fig:Colorspace}
\end{figure}

A threshold value ($\uptau$) is fixed at the beginning. We then iterate over values of $k$ for an image at strides of $5$ and calculate the inertia at each step. The value of inertia starts at infinity and reduces as the number of clusters increase. We obtain the optimal $k$ value for an image at the point where the inertia becomes lesser than the assigned threshold value. 
For locating the ideal threshold value, we test various values empirically and conclude that when the threshold is $70$, it gives visually pleasing results. From our experimentation, we deduce that determining an optimal value for $k$ is essential; otherwise, it compromises the quality of the sketch or produces redundant results. This is depicted in Figure \ref{fig:comparision_thresh}, which compares sketches produced when the threshold values were set to $50$, $70$, and $100$ respectively.

\section{Generating Colored Sketches}
\label{section:GANs sketch}
\begin{figure*}[h]
    \centering
    \includegraphics[width=13.5cm, height=10cm]{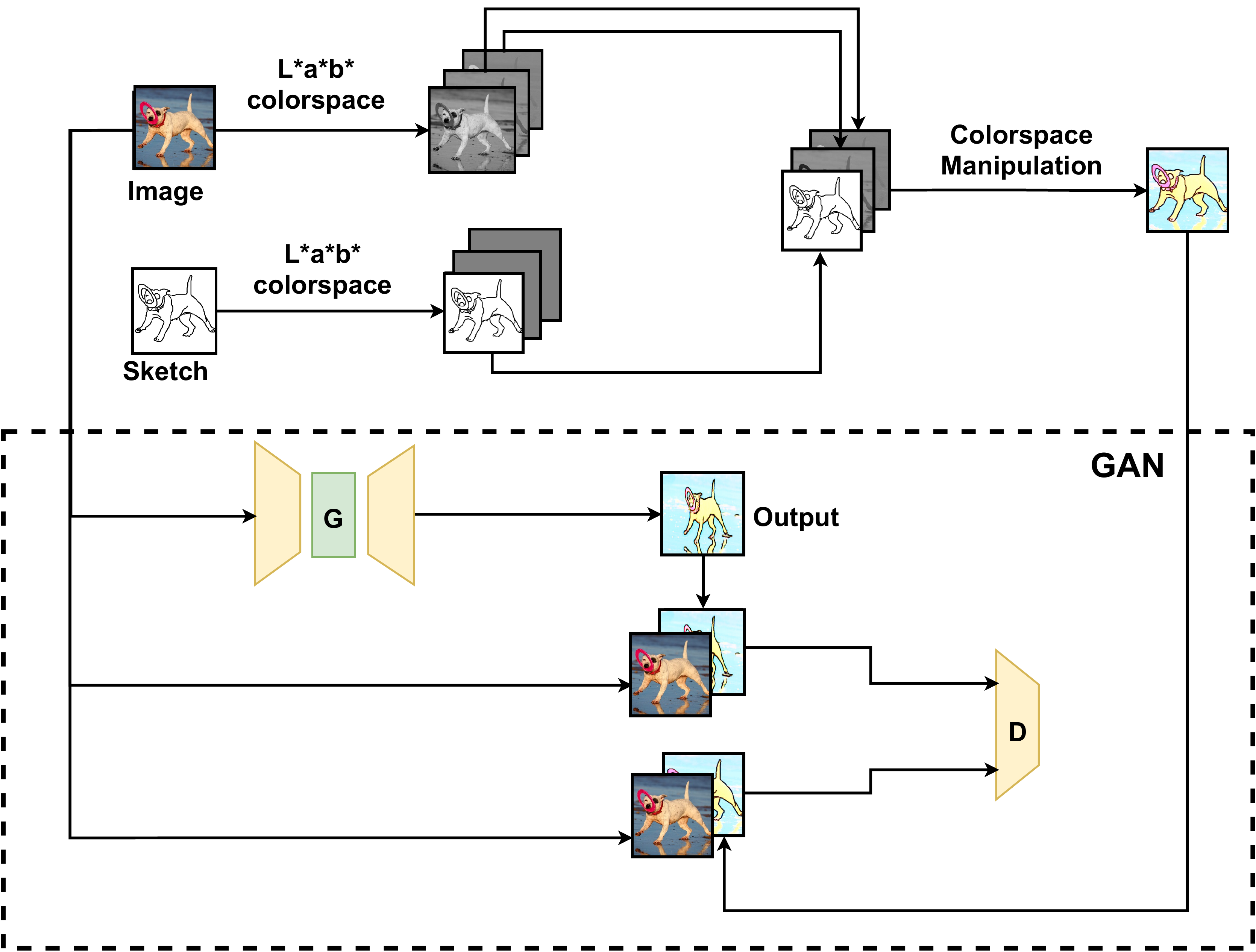}
    \vspace{6pt}
    \caption{Pipeline of our proposed generation model}
    \label{fig:GAN_pipeline}
\end{figure*}

In this section, we introduce a generative adversarial network framework that converts input images to colored sketches. We explain the colorspace manipulation administered on the dataset, illustrate the network structure, and discuss the objective functions and optimization applied on our architecture.

\subsection{Data Preprocessing}
The dataset consists of colored images and their corresponding black and white human-drawn sketches. To add color to the sketches, we propose a technique similar to the Gouache Color Transform stated in \cite{colortransform}. 

We convert the RGB image and its sketch to the $L^{*}a^{*}b^{*}$ colorspace, where $L^{*}$ stands for perceptual lightness, $a^{*}$ and $b^{*}$ for the four colors red, blue, green, yellow. The $a^{*}$ and $b^{*}$ channels of the sketch are discarded and replaced with the $a^{*}$ and $b^{*}$ channels of the corresponding image to transfer color without changing the content. The resultant image is converted back to RGB.

To improve the quality of the transferred colors, we increase the saturation by a factor of $1.8$ in HSV colorspace. We choose this factor by empirically trying different values. A value lower than this results in poor contrast, and a value higher than this makes the brighter tones appear white. The results of the preprocessing method are depicted in Figure \ref{fig:Colorspace}. These newly obtained colored sketches are included in the training data for our GAN model.

\subsection{Methodology}
GANs consist of two parts, a generator $G$ and a discriminator $D$. Our GAN learns a mapping from an input image $x$ to an output colored sketch $y$, so that $G : x\rightarrow y$. We adapt our generator and discriminator from the pix2pix work of Isola \cite{pix2pix}, where the generator resembles the architecture of U-Net \cite{u-net}, which consists of an encoder and decoder network with skip connections. The discriminator is a PatchGAN which penalizes the structure at the scale of local image patches. The entire pipeline, including the data preprocessing, is depicted in Figure \ref{fig:GAN_pipeline}.
\begin{figure}[h]
    \centering
    \includegraphics[scale=0.45]{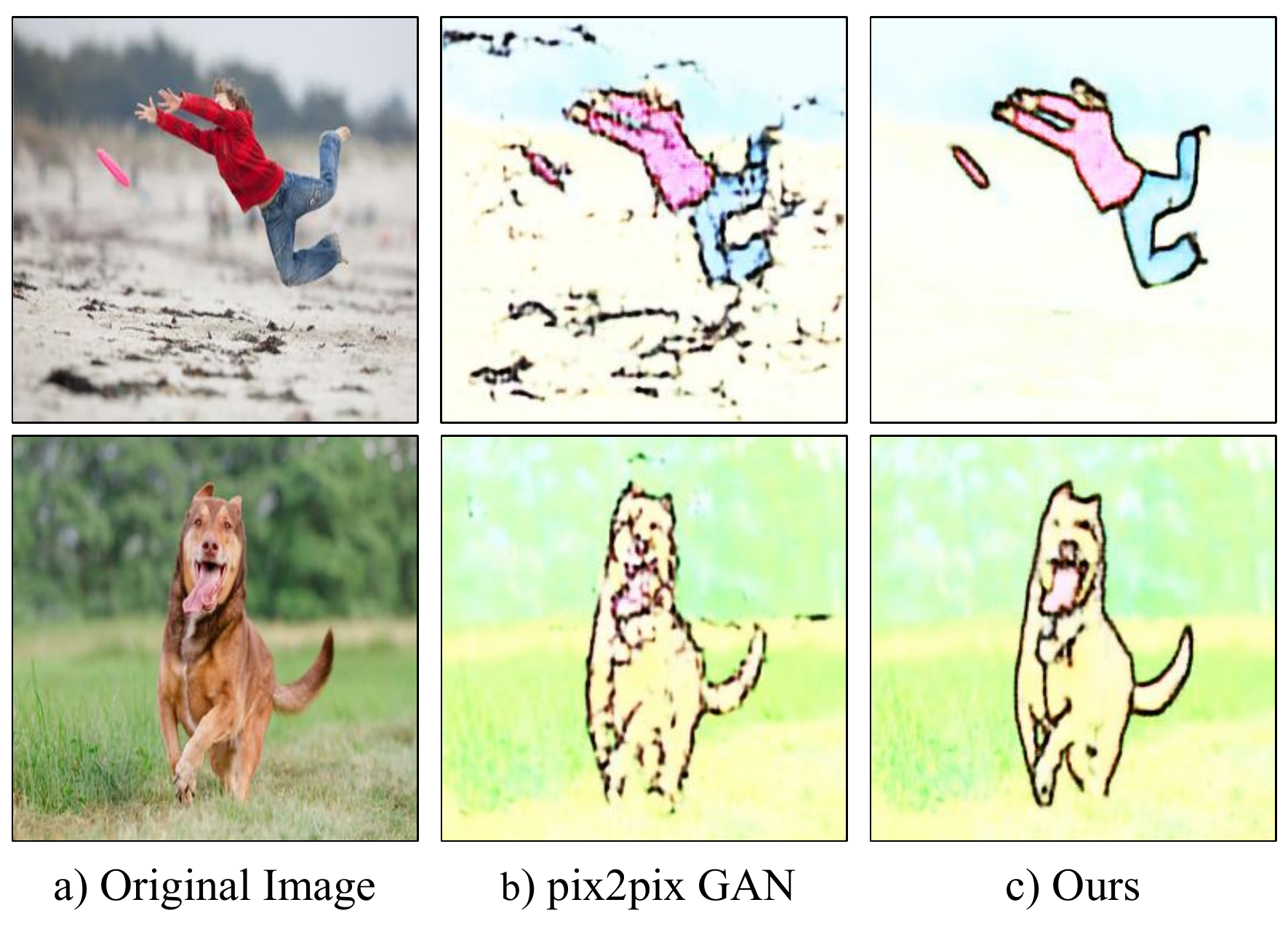}
    \caption{A visual comparision of colored sketches generated using pix2pix GAN and our GAN. Our GAN produces well-defined outlines with negligible noise.}
    \label{fig:gan_comparision}
\end{figure}

We optimize the GAN with additional convolutional layers where we use a kernel size of $3\times3$, apply same padding, and set the stride to $1$. This is followed by batch normalization and a leaky ReLU activation function at every downsampling layer of the encoder. The feature space is also increased so that more attributes can be extracted. We apply similar optimizations in the discriminator and introduce max-pooling layers alongside to extract necessary features which can help the discriminator differentiate between real and fake inputs. While passing input data to the discriminator, the post-processing colored sketches are considered real, while the output from the generator is regarded as fake. The original image is stacked upon the sketches before passing them to the discriminator. We train our GAN with the data resized to $256\times256$ dimensions followed by normalization with mean and standard deviation set to $0.5$.

\begin{figure}[b]
    \centering
    \includegraphics[scale=0.3238]{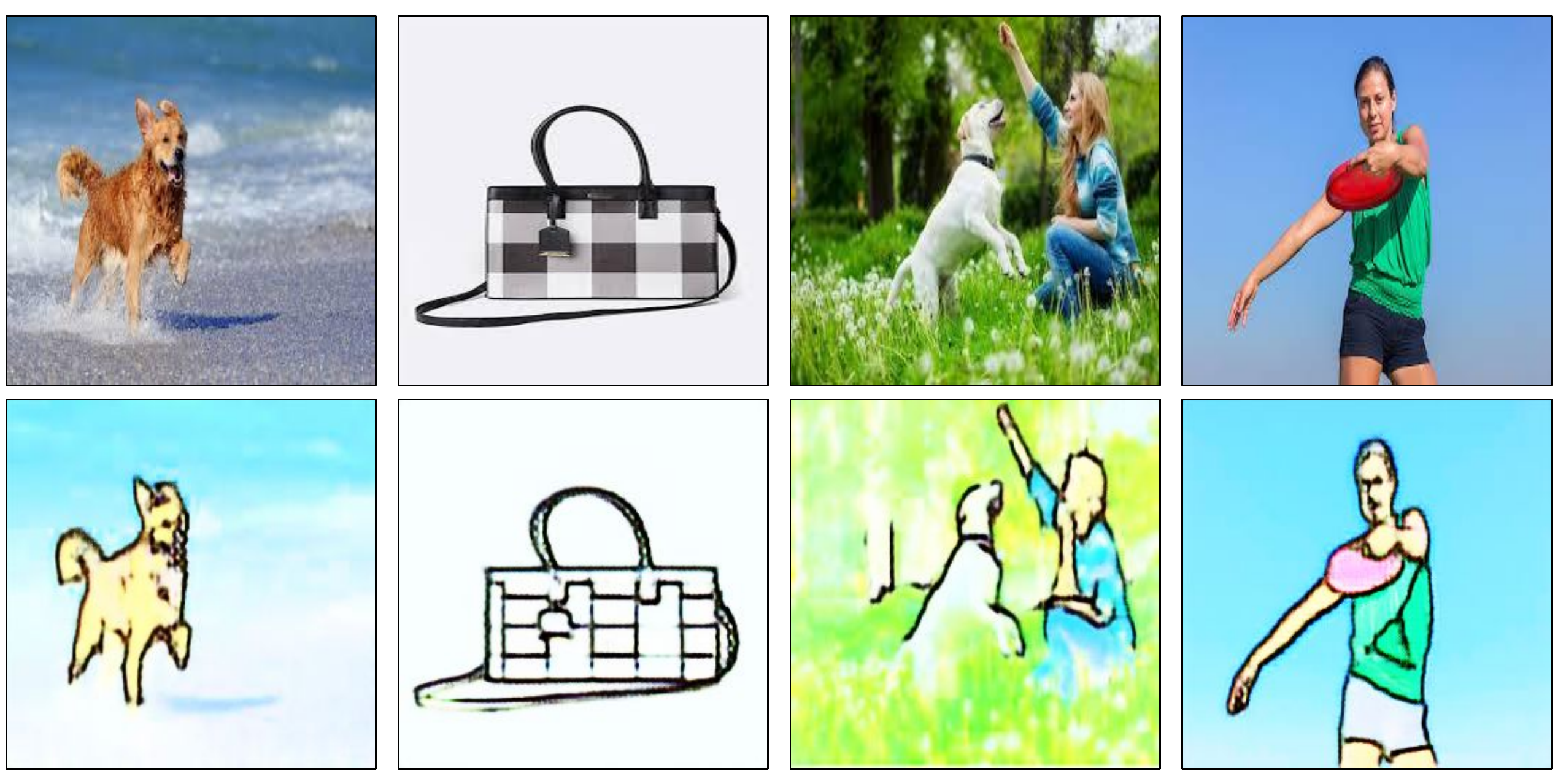}
    \caption{Sketches generated by our GAN (bottom row) from unrevealed images (top row) spanning over a variety of data distribution.}
    \label{fig:gan_test}
\end{figure}

We compare the results obtained from pix2pix and our GAN in Figure \ref{fig:gan_comparision}. The sketches generated from our model have visibly lesser noise and well-defined outlines. Our model performs better at retaining the content of the original image. Figure \ref{fig:gan_test} shows the colored sketches produced by our model from unobserved image inputs.

\subsection{Objective function and Optimization}

In a generative adversarial network \cite{gan}, a random noise vector $z$ is given as an input to the generator, which generates an output $\hat{y}=G\left ( z \right )$. The discriminator's work is to predict $y$ as real and $\hat{y}$ as fake. The loss function is defined as
\begin{equation}
\begin{split}
        L_{GAN}\left(G,D\right)=\mathop{\mathbb{E}_{y}}\left [ \log D\left (y \right ) \right ]+\\\mathop{\mathbb{E}_{z}}\left [ \log\left ( 1-  D\left ( G\left ( z \right ) \right )\right ) \right ]
\end{split}
\label{eqn:GAN}
\end{equation}

In a conditional GAN (cGAN), both the image $x$ and random noise vector $z$ are fed into the generator, which maps it to the output $\hat{y}=G\left ( x,z \right )$. The generator $G$ aims to output a sketch which resembles $y$, conditioned on $x$, while discriminator $D$ is adversarially trained to distinguish between the label $y$ and generated output $\hat{y}$ given $x$, and gives feedback to the generator on whether it is real or fake. The loss for this objective can be written as
\begin{equation}
\begin{split}
        L_{cGAN}\left(G,D\right)=\mathop{\mathbb{E}_{x,y}}\left [ \log D\left ( x,y \right ) \right ]+\\\mathop{\mathbb{E}_{x,z}}\left [ \log\left ( 1-  D\left ( x,G\left (x, z \right ) \right )\right ) \right ]
\end{split}
\label{eqn:cGAN}
\end{equation}
Usually noise vector $z$ is ignored in the optimization as found in the previous work \cite{pix2pix,photosketch_cmu,deep_mul}. We do not include $z$ in our loss function either.

We include the L1 loss, which suppresses the irrelevant details while generating the outlines for colored sketches and reduces the blurring of the output. The L1 loss is as follows
\begin{equation}
    L_{1}\left(G\right) = \mathop{\mathbb{E}_{x,y}}\left [ \left \| y-G\left(x \right) \right \|_{1} \right ]
\label{eqn:L1loss}
\end{equation}

The combined loss function now becomes
\begin{equation}
    L_{c} = \arg\min_{G}\max_{D}L_{cGAN} \left (G,D\right ) + \lambda L_{1}\left (G \right )
\label{eqn:LC}
\end{equation}
where $\lambda$ is a constant non-negative real number which controls the strength of L1 loss.

To optimize our objective function, we follow the standard approach as mentioned in \cite{pix2pix}, where we oscillate between one step of gradient descent on $D$ and one step of gradient descent on $G$. The output of the discriminator is a patch of dimensions $26\times26\times1$, which is passed to the objective function. The objective function is then divided by the batch size, which is $32$ in our case. We used Adam optimizer with a learning rate of $0.0005$. The momentum parameters are defined as $\beta_1 = 0.5$, $\beta_2 = 0.999$, and $\lambda$ is set to $1000$. 

\section{Results and Evaluation}
\label{section:results}
\paragraph{Quantitative Evaluation:} Evaluating artworks is highly subjective, thus making quantitative image assessment a challenging task. In this paper, we have used Neural Image Assessment (NIMA) \cite{NIMA} to assess our results. NIMA consists of two convolutional network models which use pre-trained weights of ImageNet classification task and are fine-tuned on Aesthetic Visual Analysis (AVA) \cite{AVA} and Tampere Image Database (TID2013) \cite{TID2013} datasets to predict the aesthetic and technical score, respectively. The technical score is a measure of low-level characteristics such as noise, blur, and saturation. The aesthetic score is used to assess semantic level characteristics such as style. The models provide the score on a scale of $1$ to $10$ where $1$ is the lowest, and $10$ is the perfect score.

We use the two NIMA models to predict the technical and aesthetic scores of our results: colored outlined sketches and colored sketches and compare them with the Contour Drawing Dataset. Additionally, we also calculate the scores on $1000$ randomly sampled sketches from three widely used sketch datasets: TU-Berlin \cite{tu-berlin}, The Sketchy Database \cite{sketchydatabase} and SketchyScene \cite{SketchyScene}. The average NIMA scores of all the sampled sketches in each dataset are listed in Table \ref{tab:nimatable}.
\renewcommand{\arraystretch}{1.3}
\begin{table}[h]
\centering
\begin{tabular}{ M{3.8cm} M{1.5cm} M{1.5cm} } 
\hline
Sketch Dataset & Technical Score & Aesthetic Score \\
\hline
TU-Berlin\cite{tu-berlin}  & \textbf{5.24} & 4.32 \\ 
The Sketchy Database\cite{sketchydatabase} & 4.83 & \textbf{5.15} \\ 
SketchyScene\cite{SketchyScene} & 4.36 & \textbf{5.15} \\ 
\hline
Contour Drawing Dataset\cite{photosketch_cmu} & 4.26 & 5.12 \\
\hline
Colored Outline (Ours) & 4.34 & 5.02 \\
Colored Sketch (Ours) & 4.24 & 4.92 \\
\hline
\end{tabular}
    \vspace{8pt}
    \caption{Comparison of NIMA technical and aesthetic scores of our results with other sketch datasets.}
    \label{tab:nimatable}
\end{table}

The colored outlines have a marginally higher average technical score than the original black and white sketches. The average technical and aesthetic scores of the output images from both methods are very close to the scores of the original sketches and the three commonly used sketch datasets.

\begin{figure}[t]
    \centering
    \includegraphics[scale=0.445]{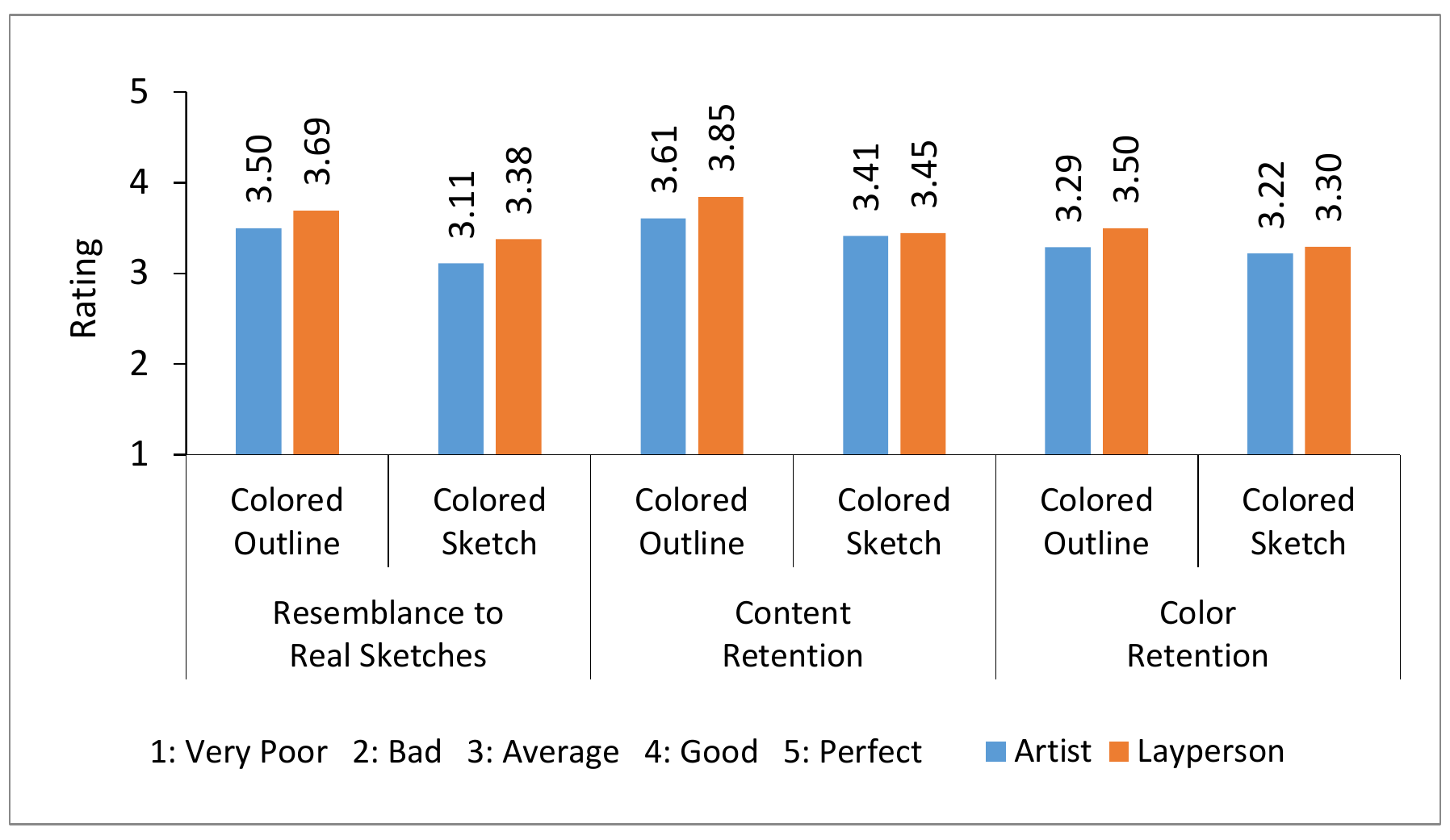}
    \caption{Results of our perceptual user study for each category rated on a scale of 1 to 5, collected from artists and laypeople.}
    \label{fig:barchart}
\end{figure}

\paragraph{Perceptual Study:} Besides evaluating the results based on NIMA scores, we also conduct a perceptual study. A user interface has been created which randomly generates a set of $10$ photographs, their corresponding colored outline sketches, and color-filled sketches at a time. We divide the users into two categories. For our purpose, we consider painters, pencil and pen artists, digital artists, illustrators, animators, and professionals from related backgrounds to be more critical of our results and identify them as artists. Others are classified as laypeople. The users are expected to view a minimum of $50$ results, and there is no maximum limit. They are then asked to rate our results on a scale of $1$ to $5$ based on three criteria: resemblance to real or human-drawn sketches, content retention in the sketches compared to the original image, and color retention in the sketches compared to the original image. We define a legend for the scores to remove ambiguity for the users as stated in Figure \ref{fig:barchart}.
A total of 100 verified users\textemdash 30 artists and 70 laypeople\textemdash voluntarily participated in our user-based evaluation process. We take the weighted average of the ratings based on the number of results viewed by each user. The results have been presented in Figure \ref{fig:barchart} which confirms that both groups of users find our results to be above average rating.

\section{Applications}
\label{section:applications}
There is a wide range of applications for our methods, as color plays a major role in art. Colored sketches are able to convey more information and details, while maintaining their sparsity. For example, a black and white sketch of a carrot has a higher chance of it being confused with a radish or a similar object. But the colored version will be recognized with more ease and less confusion.

Pre-primary schools could be one use for generating colored outlines and sketches. During this stage, students begin to learn how to sketch. The following is a description of the learning process. They begin by drawing black and white outlines and then progress to colored outlines and later to color-filled sketches. They may acquire the approach mentioned above from their tutor or duplicate it from photos they see. The teacher can choose colored images and draw the black and white outlines using the sketch game mentioned in the paper \cite{photosketch_cmu}. The teacher can feed the colored image and black \& white outlines to our model to produce the colored outlines, which can assist the students by providing a reference.

\section{Conclusion and Future Work}
\label{section:conclusion}
In this work, we examine the problem of producing colored sketches from images. We extend the Contour Drawing Dataset with two kinds of colored versions of the sketches, introduce a generation model, and compare our work with the existing literature. Our results hold a plethora of applications as they can be further annotated and utilized on tasks concerning sketch recognition, segmentation, text-based sketch generation, and sketch-based modeling. The generation model automates the process of drawing a sketch inspired by a visual scene or photograph, thus saving time and labor.

We make use of unsupervised learning and image processing techniques to render colored outlines. This can also be attempted with deep learning approaches like neural style transfer. The generation model can be improved by training on larger datasets that include a variety of objects. Our work can be extended to encompass these potential enhancements. 

{\small
\bibliographystyle{plain}
\bibliography{xcisketch}
}

\end{document}